\documentclass{article}
\usepackage{spconf,amsmath,graphicx}

\usepackage{float}
\usepackage{subfig}
\usepackage{algorithm}
\usepackage{algorithmic}
\usepackage{amsthm}
\usepackage{amssymb}
\usepackage{booktabs}

\usepackage{newfloat}
\usepackage{listings}

\usepackage[hyphens]{url}

\title{One-Step Late Fusion Multi-view Clustering with Compressed Subspace}
\name{Qiyuan Ou$^{1}$, Pei Zhang$^{1}$, Sihang Zhou$^{2}$, En Zhu$^{*1}$
	\thanks{
		This work is supported by National Key R\&D Program of China (No. 2022ZD0209103), National Natural Science Foundation of China (No. 62325604, 62276271) and Hunan Provincial Graduate Student Research Program (No.CX20230050).
	}}
\address{
	$^1$College of Computer, National University of Defense Technology\\
	$^2$College of Intelligence Science and Technology, National University of Defense Technology\\
	$^*$Corresponding author\\
	\{ouqiyuan14, enzhu\}@nudt.edu.cn	
}

\begin{document}
\ninept
\maketitle
\begin{abstract}
Late fusion multi-view clustering (LFMVC) has become a rapidly growing class of methods in the multi-view clustering (MVC) field, owing to its excellent computational speed and clustering performance. One bottleneck faced by existing late fusion methods is that they are usually aligned to the average kernel function, which makes the clustering performance highly dependent on the quality of datasets. Another problem is that they require subsequent k-means clustering after obtaining the consensus partition matrix to get the final discrete labels, and the resulting separation of the label learning and cluster structure optimization processes limits the integrity of these models. To address the above issues, we propose an integrated framework named One-Step Late Fusion Multi-view Clustering with Compressed Subspace (OS-LFMVC-CS). Specifically, we use the consensus subspace to align the partition matrix while optimizing the partition fusion, and utilize the fused partition matrix to guide the learning of discrete labels. A six-step iterative optimization approach with verified convergence is proposed. Sufficient experiments on multiple datasets validate the effectiveness and efficiency of our proposed method.
\end{abstract}
\section{Introduction}
\label{sec:intro}
\vspace*{-2mm}
The $k$-means algorithm, as a classical and widely used clustering algorithm, provides an intuitive and effective method for cluster analysis.
Denote data matrix $\mathbf{X}\in \mathbb{R}^{n\times d}$, of which each line represents an element from the set of data samples $\left\{\mathbf{x}_i\right\}_{i=1}^n \subseteq \mathcal{X}$. Denote $\mathbf{F}\in \mathbb{R}^{n\times \mu}$ as cluster indicating matrix and $\mu$ represents total number of clusters. $\mathbf{F}_{i,j}= \frac{1}{\sqrt{|C_{j}|}}$ $\Longleftrightarrow$ $x_i$ belongs to $j$-th cluster, or it equals $0$. Discrete $k$-means methods can be expressed as:
\vspace*{-3mm}
\begin{equation}
	\begin{gathered}
		\max _{\mathbf{F}} \operatorname{Tr}\left(\mathbf{F}^{\top} \mathbf{X} \mathbf{X}^{\top} \mathbf{F}\right), \\
		\text { s.t. } \mathbf{F} \in \mathbb{R}^{n \times \mu}, \mathbf{F}_{i j}=\left\{\begin{array}{l}
			\frac{1}{\sqrt{\left|\mathbf{C}_j\right|}}, \text { if } x_i \text { is in the } j \text {-th cluster. } \\
			0, \text { otherwise. }
		\end{array}\right.
	\end{gathered}
\end{equation}

To optimize the above target function concerns NP-hard problem. Thereby traditional $k$-means clustering algorithms relax the discrete constraints on $\mathbf{F}$ matrix to orthogonal constraints. The terms we use in this paper are listed in Table\ref{notations}.

\vspace*{-2mm}
\begin{table}[H]\small
	\centering
	\renewcommand\arraystretch{0.86}
	\tabcolsep=0.28cm
	\caption{NOMENCLATURE}\label{notations}\vspace*{-1mm}
	\begin{tabular}{c|c}
		\toprule
		$p$ & The number of views. \tabularnewline
		$\mu$ & The number of clusters. \tabularnewline
		$k$ & The dimension of partition matrices. \tabularnewline
		$m$ & The scale of compressed subspace. \tabularnewline
		$\mathbf{I}$ & Identity matrix. \tabularnewline
		$\mathbf{X}\in \mathbb{R}^{n\times d}$ & A series of $n$ samples with $d$ dimensions. \tabularnewline
		$\|\cdot\|_{\mathrm{F}}$ & Frobenius norm. \tabularnewline
		$\boldsymbol{\beta}\in \mathbb{R}^{p}$ &A concatenation of $p$ weights. \tabularnewline
		$\left\{\mathbf{K}_i\right\}_{i=1}^p$ & The $i$-th base kernel matrix. \tabularnewline
		$\mathbf{Y}\in \{0,1\}^{\mu \times n}$ & Label matrix. \tabularnewline
		$\mathbf{C}\in\mathbb{R}^{k \times \mu}$ & The clustering centroids. \tabularnewline
		$\left\{\mathbf{W}_i\right\}_{i=1}^p$ & The permutation matrix of $i$-th partition. \tabularnewline
		$\mathbf{H} \in \mathbb{R}^{k \times n}$ & Partition matrix of consensus embedding. \tabularnewline
		$\mathbf{H}_i \in \mathbb{R}^{k \times n}$ & The partition matrix of individual kernels. \tabularnewline
		$\mathbf{P} \in \mathbb{R}^{n \times m}$ & The unified compression matrix. \tabularnewline
		$\mathbf{S} \in \mathbb{R}^{m \times n}$ & The consensus reconstruction matrix.\tabularnewline
		\bottomrule
	\end{tabular}
\end{table}
\vspace*{-2mm}
For datasets that are not linearly separable, kernel mappings are utilized to perform kernel $k$-means algorithms\cite{dhillon2004kernel,chitta2011approximate}, which can be easily adopted in multi-view clustering tasks\cite{du2015robust,zhang2020consensus,liu2022simplemkkm}. Denote $ \phi _{i}: \mathcal{X} \rightarrow \mathcal{H} _{i}$ as the $i$-th feature mapping from $ \left\{ \mathbf{x}_{i}\right\} _{i=1}^{n}$ to $p$ Regenerative Kernel Hilbert Space (RKHS) $\left\{{\mathcal{H}}_i\right\}_{i=1}^p$. Each sample in multiple kernel clustering is denoted as $ \phi _{\boldsymbol{\beta}}(\mathbf{x})= \left[ \beta _{1}\phi _{1}(\mathbf{x})^{\top}, \cdots , \beta _{p} \phi _{p}(\mathbf{x})^{\top}\right] ^{\top}$, where ${\boldsymbol{\beta}}$ is the coefficient vector of $p$ base kernel functions. The kernel weights are adjusted during the clustering process to optimize the clustering performance. 
Kernel function can be expressed by:
	$\kappa_{\boldsymbol{\beta}}\left(\mathbf{x}_i, \mathbf{x}_j\right)=\phi_{\boldsymbol{\beta}}\left(\mathbf{x}_i\right)^{\top} \phi_{\boldsymbol{\beta}}\left(\mathbf{x}_j\right)$,
and the corresponding loss function of Multiple Kernel K-means (MKKM) is:
\vspace*{-2mm}
\begin{equation}\label{mkkm}
	\begin{array}{ll}
		\min\limits_{\mathbf{H}, \boldsymbol{\beta}} & \operatorname{Tr}\left(\mathbf{K}_{\boldsymbol{\beta}}\left(\mathbf{I}_n-\mathbf{H}^{\top}\mathbf{H}\right)\right), \\
		\text { s.t. } & \mathbf{H} \in \mathbb{R}^{k \times n},  \mathbf{H}\mathbf{H}^{\top}=\mathbf{I}_k, \\
		& \boldsymbol{\beta}^{\top} \mathbf{1}_p=1, \boldsymbol{\beta} \geq 0.
	\end{array}
\end{equation}
Optimizing the objective can be reduced to a traditional $k$-means process and solving a quadratic programming problem through alternately optimizing $\boldsymbol{\beta}$ and $\mathbf{H}$.

There are plenty of work emerged to improve MKKM, e.g., linear fusion-based methods\cite{conf/aaai/LiuDY0Z16} assumes that each base kernel function can correspond to a view of different dimensions, which in turn captures complementary aspects of features, and the multi-view consistent kernel matrices can be obtained by linear combination of multiple base kernel matrices. Secondly, joint training-based methods\cite{ren2021multiple,lu2022scalable} assumes that the clustering results can be obtained independently based on each base kernel matrix, and that the multi-view clustering results should be consistent, so the clustering information can be fused by statistical methods to further enhance the credibility of the clustering results. Thirdly, consensus structure extraction methods\cite{liu2020cluster,lu2022multiple} assumes that different kernel functions acting on the data matrix can preserve and extract the consistent clustering structure existing in multiple types of data, and the consistent clustering structure of multiple kernels is obtained by decomposing the kernel matrix. There are also deep learning based methods adopted to improve clustering performance and learn representations\cite{liuyue_Dink_net,wang2022test,wang2022localized,2021DFCN,2023RARE,TGC_ML,2022ResCapsNet,2023CMSCGC}. To improve the efficiency and to better represent the sample distributions of individual views, late fusion based multi-kernel clustering methods\cite{DBLP:conf/ijcai/WangLZTLHXY19,wang2021late} use a tighter approach (base partitioning matrix $\mathbf{H}\in \mathbb{R}^{k\times n}$ extracted using $k$-means algorithm) for structural representation rather than using multiple base kernels.

We propose a One-Step Late Fusion Multi-view Clustering with Compressed Subspace method to directly obtain discrete cluster label by integrating clustering structure optimization and label learning into a unified framework. Our proposed algorithm has the following advantages: 
\begin{itemize}
	\item Our algorithm is able to obtain clustering labels in one step, by negotiating label learning and cluster structure optimization through a unified framework. \vspace{-1mm}
	\item The method is highly efficient with both $\mathcal{O}(n)$ time and space expenditure, which allows our algorithm to be used directly on large-scale multi-view datasets.\vspace{-1mm}
	\item We propose a six-step iterative optimization algorithm with fast convergence of the target. We conduct experiments to verify the effectiveness and efficiency of the algorithm.\vspace{-1mm}
\end{itemize}\vspace{-1mm}

\section{Methodology}
\label{sec:method}

Most existing multi-kernel clustering methods assume that the optimal kernel exists in a linear space consisting of base kernels\cite{conf/aaai/LiuDY0Z16}, and such assumption greatly limits the feasible domain of the optimal kernel. ONKC methods\cite{DBLP:journals/tkde/LiuLXLZWY22} propose to reconstruct the kernel matrices in a nonlinear neighborhood space, thus enlarging the search space of the optimal kernel, but the computational overheads of such methods are $\mathcal{O}(n^3)$, which restrict the algorithms' application to large-scale datasets. LFMVC methods\cite{wang2019efficient,wang2021late} reduce the dimensionality of the kernel matrix by constructing the corresponding base partition matrices, thus reducing the computational overhead. These methods usually use the average kernel as a reference for the alignment of partitions, and requires higher quality of these base partition matrices. These drawbacks limit their generalizability to the clustering task over wide range of datasets. 

In addition, the above methods cannot directly obtain cluster labels and require an extra spectral clustering or $k$-means process for ultimate clusters. To tackle the NP-hard problem brought by discrete cluster labels, spectral rotation (SR)\cite{conf/aaai/HuangNH13a,journals/tcyb/PangXNL20,DBLP:conf/icassp/LuLWN22} and improved spectral rotation (ISR)\cite{conf/ijcai/0004NHY17} methods are proposed which learn the discrete labels and representations synchronically. However, the $\mathcal{O}(n^3)$ time cost and $\mathcal{O}(n^2)$ space cost greatly inhibits the scalability for large-scale datasets.


We adopt kernel subspace clustering method\cite{DBLP:conf/nips/HamL08} for self-reconstruction of the consensus kernel partition, and use the trace alignment rather than Frobenius norm to avoid a re-weighing procedure\cite{DBLP:conf/icml/NieYH14}. We additionally adopt a compressed subspace using a uniform matrix $\mathbf{P}$ to further increase the computational efficiency. It is proved by\cite{journals/pami/LiuZLWTYSWG19} that maximizing late-fusion alignment is equivalent to minimizing the target function of MKKM in Eq.\ref{mkkm}. By integrating late fusion of kernel partition alignment maximization and self-reconstruction through a shared reconstruction matrix $\mathbf{S}$, our method is capable of optimizing cluster structure with the negotiation of multiple views. A cluster label assigning matrix $\mathbf{Y}$ is learned with the centroid $\mathbf{C}$ to refine the aligned partition $\mathbf{H}$. The overall optimization target is listed below:
\vspace*{-6mm}
\begin{center}
	\begin{equation}\label{target2}
		\begin{aligned}
			& \max _{\mathbf{P}, \mathbf{S},\left\{\mathbf{W}_i\right\}_{i=1}^p, \boldsymbol{\beta},\mathbf{C},\mathbf{Y}}\mathrm{tr}(\mathbf{H}(\mathbf{H P S})^{\top}+ {\mathbf{Y}}^{\top}{\mathbf{C}}^{\top}\mathbf{H}) \\
			\text { s.t. } &\mathbf{P}^{\top} \mathbf{P}=\mathbf{I}_m, \mathbf{S} \geq 0, \sum_{i=1}^m\mathbf{S}_{i,j}^2 = 1,\forall j \in \{1,2,\cdots,n\}, \\
			&  \boldsymbol{\beta}_i \geq 0, \forall i \in \{1,2,\cdots,p\},\boldsymbol{1}_{\mu}^{\top}\mathbf{Y}=\boldsymbol{1}_{n}^{\top},\mathbf{Y}\in \{0,1\}^{\mu \times n},\\
			&  \mathbf{H}=\sum_{i=1}^p \boldsymbol{\beta}_i \mathbf{W}_i \mathbf{H}_i,\mathbf{W}_i\mathbf{W}_i^{\top} =\mathbf{I}_k,\mathbf{C}^{\top}\mathbf{C}=\mathbf{I}_{\mu},			
		\end{aligned}
	\end{equation}
\end{center}\vspace{-2mm}
where $\boldsymbol{1}_{n}\in\mathbb{R}^{n}$ is an all-$1$ column vector.
\vspace{-2mm}

\section{Optimization}
\label{sec:optim}

We develop a six-step iterative optimization algorithm to maximize the clustering target \ref{target2}. 

\noindent{\textbf{Update $\left\{{\mathbf{W}}_i\right\}_{i=1}^p $ with $\mathbf{P}$, $\mathbf{S}$, $\boldsymbol{\beta}$, $\mathbf{C,Y}$ fixed.}} For $\forall \delta \in\{1,2, \cdots, p\}$, we update $\mathbf{W}_{\delta}$ by maximizing the target listed below:
\vspace{-2mm}
\begin{equation}\label{op_W2}
	\begin{aligned}
		\max _{{\mathbf{W}}_\delta} \operatorname{tr}\left(\mathbf{G} {\mathbf{W}}_\delta^{\top}\right) 
		~\text{s.t.}{\mathbf{W}}_\delta^{\top} {\mathbf{W}}_\delta=\mathbf{I}_k,
	\end{aligned}
\end{equation}
where $\mathbf{G}=\beta_{\delta}\left(\sum_{j=1, j \neq \delta}^p \beta_j {\mathbf{W}}_j \mathbf{H}_j\right) \mathbf{S}^{\top} \mathbf{P}^{\top} \mathbf{H}_\delta^{\top} + \beta_{\delta}\mathbf{CYH}_{\delta}^{\top}=\mathbf{U}_g \mathbf{D}_g \mathbf{V}_g^{\top}$, $\mathbf{U}_g$ and $\mathbf{V}_g$ are the left and right singular matrices, respectively. According to \cite{ou2020anchor}, a closed-form solution of the optimal can be expressed as:
\vspace{-2mm}
\begin{equation}\label{op_W3}
	{\mathbf{W}}_\delta^*=\mathbf{U}_g^* \mathbf{V}_g^{\top}.
\end{equation}
By following the updating formula in Eq. \eqref{op_W3}, the algorithm refreshes $\mathbf{W}_i,~i=1,2,\cdots,p$ in succession.

\noindent{\textbf{Update $\boldsymbol{\beta}$ with $\left\{{\mathbf{W}}_i\right\}_{i=1}^p$, $\mathbf{P}$, $\mathbf{S}$, $\mathbf{C,Y}$ fixed.}}
When $\mathbf{P}$, $\mathbf{S}$, $\mathbf{C,Y}$ are settled, updating $\boldsymbol{\beta}$ with fixed $\left\{{\mathbf{W}}_i\right\}_{i=1}^p$, which concludes the final fusion step, can be termed as: 
\vspace*{-8mm}
\begin{center}
	\begin{equation}\label{op_beta1}
		\begin{aligned}
			 &\max _{\boldsymbol{\beta}} \sum_{i=1}^p \sum_{j=1}^p \beta_i \beta_j \operatorname{tr}\left(\mathbf{P} \mathbf{S} \mathbf{H}_i^{\top} \mathbf{W}_i^{\top} \mathbf{W}_j \mathbf{H}_j\right) \\
			 &+ \sum_{i=1}^p \beta_i \operatorname{tr}\left(\mathbf{C}^{\top}\mathbf{W}_i\mathbf{H}_i\mathbf{Y}^{\top}\right)
			 \quad \text { s.t. } {\boldsymbol{\beta}} \geqslant 0,
		\end{aligned}
	\end{equation}
\end{center}\vspace*{-3mm}
denoting the quadratic coefficient matrix $\mathbf{M}$ where $\mathbf{M}_{i,j} ~= -\operatorname{tr}\left(\mathbf{P} \mathbf{S} \mathbf{H}_i^{\top} \mathbf{W}_i^{\top} \mathbf{W}_j \mathbf{H}_j\right)$, and coefficient vector $f$ with $f_i=-\operatorname{tr}\left(\mathbf{C}^{\top}\mathbf{W}_i\mathbf{H}_i\mathbf{Y}^{\top}\right)$, it can be further written as a minimization function:
\vspace*{-1mm}
\begin{center}
	\begin{equation}\label{op_beta2}
		\begin{aligned}
			\min _{\boldsymbol{\beta}} {\boldsymbol{\beta}}^T \mathbf{M} {\boldsymbol{\beta}} + f{\boldsymbol{\beta}}^T \text {, s.t. } {\boldsymbol{\beta}} \geqslant 0.
		\end{aligned}
	\end{equation}
\end{center}
It is worth noticing that Eq. \eqref{op_beta2} is a quadratic optimization problem and a symmetrization manipulation with respect to $\mathbf{M}$ does not affect the results of the quadratic objective equation. Alternatively it can acquire a closed-form solution by Cauchy-Schwarz Inequality after diagonalization of the polarized $\mathbf{M}$. Here we directly adopt the quadratic programming scheme\cite{DBLP:journals/siamjo/ColemanL96a} to achieve higher speed.

\noindent{\textbf{Update $\mathbf{P}$ with $\boldsymbol{\beta}$, $\left\{{\mathbf{W}}_i\right\}_{i=1}^p$, $\mathbf{S}$, $\mathbf{C,Y}$ fixed.}}
The optimization problem in Eq. \eqref{target2} concerning $\mathbf{P}$ can be written as:
\vspace*{-10mm}
\begin{center}
	\begin{equation}\label{op_P1}
		\vspace{-5pt}
		\begin{aligned}
			& \max _\mathbf{P} \operatorname{tr}\left(\mathbf{H}(\mathbf{H} \mathbf{P} \mathbf{S})^{\top}\right) \quad \text { s.t. } \mathbf{P}^{\top} \mathbf{P}=\mathbf{I}_m, \mathbf{H}=\sum_{i=1}^p \beta_i \mathbf{W}_i \mathbf{H}_i 
		\end{aligned}
	\end{equation}
\end{center}

Likewise, the deduction for the optimization of the compress matrix $\mathbf{P}$ is similar to Eq. \eqref{op_W2}. Denote $\mathbf{A}=\mathbf{H}^{\top} \mathbf{H} \mathbf{S}=\mathbf{U}_a \mathbf{D}_a \mathbf{V}_a^{\top}$, the optimal solution is:
\vspace*{-2mm}
\begin{equation}\label{op_P2}
	\mathbf{P}^*=\mathbf{U}_a \mathbf{V}_a^{\top}.
\end{equation}

\vspace{-2mm}
\noindent{\textbf{Update $\mathbf{S}$ with $\boldsymbol{\beta}$, $\left\{{\mathbf{W}}_i\right\}_{i=1}^p$, $\mathbf{P}$, $\mathbf{C,Y}$ fixed.}}
When variables $\boldsymbol{\beta}$, $\left\{{\mathbf{W}}_i\right\}_{i=1}^p$ are fixed, given compress matrix $\mathbf{P}$, the late-fusion kernel is lodged in a unified compressed subspace, which is shared among views. Updating $\mathbf{S}$ is equivalent to constructing the bipartite graph from the anchor space to the original feature space. However in our method, the consensus subspace does not need further integration.
The optimization procedure regarding $\mathbf{S}$ is written below:
\vspace*{-5mm}
\begin{center}
	\begin{equation}\label{op_S1}
		\begin{aligned}
			& \max _\mathbf{S} \operatorname{tr}\left(\mathbf{H}^{\top} \mathbf{H} \mathbf{P} \mathbf{S}\right) \quad \text { s.t. } \sum_{i=1}^m \mathbf{S}_{i, j}^2=1, \forall_j \in\{1,2, \cdots, n\}.
		\end{aligned}
	\end{equation}
\end{center}
Let $ \mathbf{Q}=\mathbf{H}^{\top} \mathbf{H} \mathbf{P}=\left({\boldsymbol{q}}_1~{\boldsymbol{q}}_2~\cdots~{\boldsymbol{q}}_n
\right)^{\top}, \mathbf{S}=\left({\boldsymbol{s}}_1~{\boldsymbol{s}}_2~ \cdots ~ {\boldsymbol{s}}_n\right)$, where $\boldsymbol{q}$ and $\boldsymbol{s}$ are column vectors, Eq. \eqref{op_S1} is equivalent to:
\vspace{-8mm}
\begin{center}
	\begin{equation}\label{op_S2}
		\max _\mathbf{S} \sum_{i=1}^n {\boldsymbol{q}}_i {\boldsymbol{s}}_i,\quad \text { s.t. } \sum_{i=1}^m \mathbf{S}_{i, j}^2=1,
	\end{equation}
\end{center}
accordingly, the optimal solution is:\vspace{-2mm}
\begin{equation}\label{op_S3}
	\mathbf{S}_{j, i}^*=\frac{\mathbf{Q}_{i j}}{\left\|{\boldsymbol{q}}_i\right\|_2}
\end{equation}

\noindent{\textbf{Update $\mathbf{C}$ with $\boldsymbol{\beta}$, $\left\{{\mathbf{W}}_i\right\}_{i=1}^p$, $\mathbf{P}$, $\mathbf{Y}$, $\mathbf{S}$ fixed.}}
The optimization problem w.r.t. centroids $\mathbf{C}$ is:
\vspace{-1mm}
\begin{equation}
	\max _\mathbf{C} \operatorname{tr}\left(\mathbf{HY^{\top}C^{\top}}\right), s.t. \mathbf{C}^{\top}\mathbf{C}=\mathbf{I}_{\mu},
\end{equation}
likewise, denoting $\mathbf{HY^{\top}}=\mathbf{U}_c\mathbf{D}_c \mathbf{V}_c^{\top}$, the optimal solution is:
\vspace{-1mm}
\begin{equation}\label{op_C}
	\mathbf{C}^* = \mathbf{U}_c\mathbf{V}_c^{\top}
\end{equation}

\noindent{\textbf{Update $\mathbf{Y}$ with $\boldsymbol{\beta}$, $\left\{{\mathbf{W}}_i\right\}_{i=1}^p$, $\mathbf{P}$, $\mathbf{C}$, $\mathbf{S}$ fixed.}}
The optimization problem w.r.t. discrete label $\mathbf{Y}$ is:
\vspace{-2mm}
\begin{equation}
	\max _\mathbf{Y} \operatorname{tr}\left(\mathbf{(C^{\top}H)^{\top}Y}\right), s.t. \boldsymbol{1}_{\mu}^{\top}\mathbf{Y}=\boldsymbol{1}_{n}^{\top}, \mathbf{Y}\in \{0,1\}^{\mu \times n},
\end{equation}
denoting $\mathbf{B} = \mathbf{C^{\top}H}$, the optimal solution is: 
\vspace{-2mm}
\begin{equation}\label{op_Y}
	\mathbf{Y}^*(i,j) = arg \max_i (\mathbf{B}(i,j)), \forall j\in \{1,2,\cdots,n\}.
\end{equation}
The overall optimization is listed in Algorithm \ref{Algorithm_1}. According to each step of the algorithm, the time and space cost are $\mathcal{O}(n)$, and the efficiency is validated in Sec.\ref{sec:exp}.

\begin{algorithm}[htb]
	\renewcommand{\algorithmicrequire}{\textbf{Input:}}
	\renewcommand\algorithmicensure {\textbf{Output:} }
	\caption{One-Step Late Fusion Multi-view Clustering with Compressed Subspace}
	\label{Algorithm_1}
	\begin{algorithmic}[1]
		\REQUIRE ~~\\
		Multiple base kernels $\left\{\mathbf{K}_i\right\}_{i=1}^p$, number of clusters $\mu$, the scale of compressed subspace $m$, the dimension of partition matrix $k$.
		\ENSURE ~~\\
		The label matrix $\mathbf{Y}$.\\
		\STATE \textbf{Initialization} Initialize compression matrix $\mathbf{P} \in \mathbb{R}^{n \times m}$ as orthogonalization of a randomized matrix. Initialize $\mathbf{S}$ by imposing sum-$1$ restriction of $\ell 2$-norm on randomized matrix. $\beta_i=1/p, \forall i$. $\left\{\mathbf{W}_i\right\}_{i=1}^p = \mathbf{I}_k$, $t=1$.
		\REPEAT
		\STATE Calculate $\left\{\mathbf{W}_i\right\}_{i=1}^p$ by optimizing Eq.~\eqref{op_W3};
		\STATE Calculate $\boldsymbol{\beta}$ by optimizing Eq.~\eqref{op_beta2};
		\STATE Calculate $\mathbf{P}$ by optimizing Eq.~\eqref{op_P2};
		\STATE Calculate $\mathbf{S}$ by optimizing Eq.~\eqref{op_S3};
		\STATE Calculate $\mathbf{C}$ by optimizing Eq.~\eqref{op_C};
		\STATE Calculate $\mathbf{Y}$ by optimizing Eq.~\eqref{op_Y};
		\STATE $t=t+1$.
		\UNTIL{$ (obj^{(t)} - obj^{(t-1)})^2  < 10^{-3}$}.
	\end{algorithmic}
\end{algorithm}
\vspace{-1mm}

\section{Experiment}
\label{sec:exp}
\vspace{-2mm}
In this section, we conduct a series of experiments to evaluate the effectiveness and efficiency of our proposed method.
\begin{table*}[ht]
	\centering
	\renewcommand\arraystretch{0.70}
	\tabcolsep=0.28cm
	\caption{Comparison of clustering performance on 5 benchmark datasets. The best performances are in bold-face. }\label{result}
	\begin{tabular}{@{}ccccccccl@{}}
		\toprule
		Datasets    & Avg-KKM    & SB-KKM     & MKKM       & RMSC       & FMKKM               & FMR        & LSGMC      & \multicolumn{1}{c}{\textbf{Proposed}}            \\ \midrule
		\multicolumn{9}{c}{ACC}                                                                                                                                   \\ \midrule
		Citeseer    & 20.8 ± 0.0 & 46.3 ± 0.2 & 20.1 ± 0.0 & 19.9 ± 0.3 & 30.6 ± 0.7          & 23.9 ± 0.0 & 22.0 ± 0.4 & \multicolumn{1}{c}{\textbf{56.6 ± 0.0}} \\
		Cora        & 30.7 ± 0.8 & 45.2 ± 0.1 & 25.3 ± 0.4 & 20.2 ± 0.1 & 38.4 ± 0.1          & 40.7 ± 0.5 & 20.9 ± 0.2 & \multicolumn{1}{c}{\textbf{60.8 ± 0.0}} \\
		ProteinFold & 29.0 ± 1.5 & 33.8 ± 1.3 & 27.0 ± 1.1 & 31.2 ± 1.0 & 32.4 ± 1.8          & 34.5 ± 1.4 & 32.8 ± 1.2 & \multicolumn{1}{c}{\textbf{35.3 ± 0.0}} \\
		NUSWIDE     & 12.5 ± 0.4 & 12.2 ± 0.3 & 12.7 ± 0.2 & -          & 14.0 ± 0.3          & -          & -          & \multicolumn{1}{c}{\textbf{14.7 ± 0.0}} \\
		Reuters     & 45.5 ± 1.5 & 47.2 ± 0.0 & 45.4 ± 1.5 & -          & 45.5 ± 1.6          & -          & -          & \multicolumn{1}{c}{\textbf{53.7 ± 0.0}} \\ \midrule
		\multicolumn{9}{c}{NMI}                                                                                                                                   \\ \midrule
		Citeseer    & 2.3 ± 0.0  & 23.2 ± 0.5 & 1.9 ± 0.0  & 0.4 ± 0.1  & 10.1 ± 0.4          & 2.3 ± 0.0  & 1.7 ± 0.1  & \textbf{28.7 ± 0.0}                     \\
		Cora        & 15.7 ± 1.4 & 25.6 ± 0.1 & 9.5 ± 0.2  & 1.7 ± 0.1  & 21.8 ± 0.1          & 20.0 ± 0.2 & 0.7 ± 0.0  & \textbf{37.8 ± 0.0}                     \\
		ProteinFold & 40.3 ± 1.3 & 41.1 ± 1.1 & 38.0 ± 0.6 & 43.2 ± 0.8 & 41.5 ± 1.0          & 42.0 ± 1.1 & 43.9 ± 0.5 & \textbf{44.1 ± 0.0}                     \\
		NUSWIDE     & 11.1 ± 0.1 & 11.0 ± 0.1 & 11.3 ± 0.2 & -          & 12.6 ± 0.2          & -          & -          & \textbf{13.2 ± 0.0}                     \\
		Reuters     & 27.4 ± 0.4 & 25.5 ± 0.0 & 27.3 ± 0.4 & -          & 27.6 ± 0.5          & -          & -          & \textbf{31.8 ± 0.0}                     \\ \midrule
		\multicolumn{9}{c}{purity}                                                                                                                                \\ \midrule
		Citeseer    & 24.9 ± 0.0 & 48.8 ± 0.4 & 24.2 ± 0.0 & 22.1 ± 0.3 & 32.9 ± 0.7          & 25.8 ± 0.0 & 24.9 ± 0.2 & \textbf{58.9 ± 0.0}                     \\
		Cora        & 41.5 ± 1.3 & 52.5 ± 0.1 & 36.1 ± 1.0 & 31.5 ± 0.0 & 46.9 ± 0.1          & 42.9 ± 0.5 & 30.2 ± 0.0 & \textbf{63.3 ± 0.0}                     \\
		ProteinFold & 37.4 ± 1.7 & 39.4 ± 1.2 & 33.7 ± 1.1 & 38.5 ± 0.9 & 38.6 ± 1.5          & 40.6 ± 1.4 & 40.7 ± 0.6 & \textbf{43.4 ± 0.0}                     \\
		NUSWIDE     & 23.3 ± 0.3 & 23.7 ± 0.2 & 24.2 ± 0.4 & -          & \textbf{25.7 ± 0.4} & -          & -          & 25.6 ± 0.0                              \\
		Reuters     & 53.0 ± 0.4 & 53.9 ± 0.0 & 52.9 ± 0.5 & -          & 53.1 ± 0.4          & -          & -          & \textbf{61.9 ± 0.0}                     \\ \bottomrule
	\end{tabular}
\end{table*}\vspace{-2mm}
\subsection{Evaluation Preliminaries}
\vspace{-2mm}
\noindent\textbf{Datasets}
We adopt 5 publicly available multi-view benchmark datasets, including 
\emph{Citeseer}\footnote{\url{http://linqs-data.soe.ucsc.edu/public/lbc/}}, 
\emph{Cora}\footnote{\url{http://mlg.ucd.ie/aggregation/}}, \emph{ProteinFold}\footnote{\url{http://mkl.ucsd.edu/dataset/protein-fold-prediction}}, 
\emph{NUS-WIDE}\footnote{\url{https://lms.comp.nus.edu.sg/wp-content/uploads/2019/research/nuswide/NUSWIDE.html}}, 
\emph{Reuters}\footnote{\url{https://kdd.ics.uci.edu/databases/reuters21578/}}.
Among them, NUS-WIDE and Reuters are large scale datasets with over 10 thousand samples each.

\noindent\textbf{Compared Algorithms}
7 algorithms are compared with our algorithm, including multiple kernel clustering and multi-view subspace clustering methods, over the above 5 benchmark datasets. Specifically, we use \emph{SB-MKKM} and \emph{Avg-MKKM} as baseline methods of kernel clustering, which calculate the best performance and average performance of kernel $k$-means results respectively. We select \emph{MKKM}\cite{DBLP:journals/tfs/HuangCC12}, \emph{FMKKM}\cite{DBLP:conf/aaai/ZhangL0DZXZ22} as a representation of classical kernel methods which expand from fuzzy $k$-means to late-fusion-based methods. And \emph{RMSC}\cite{DBLP:conf/aaai/XiaPDY14} is representative of spectral clustering methods. In subspace clustering area, we select \emph{FMR}\cite{DBLP:conf/ijcai/LiZHZW19} and \emph{LSGMC}\cite{10089426} for performance comparison. 



\subsection{Performance Analysis}

We use the initialization specified in Algorithm \ref{Algorithm_1}. The Table \ref{result} shows the ACC, NMI and purity of all methods.  To increase the confidence of results, we conduct 20 iterations for each clustering algorithm and within $k$-means function, the replicates are set to 10. The best performance and standard variance, which is brought by $k$-means tasks, are both reported in our experiments. As our algorithm is one-step method without the downstream $k$-means or spectral clustering, the variance is 0 accordingly, guaranteeing the stability of our algorithm. The symbol `-' represents `Out-of-Memory' problem. The platform for all methods is PC with Intel(R) Core(TM) i7-12700H 2.30 GHz, 64GB RAM.

\vspace{-1mm}

\begin{figure}[htb]
	\begin{minipage}[b]{.48\linewidth}
		\centering
		\centerline{\includegraphics[width=4.0cm]{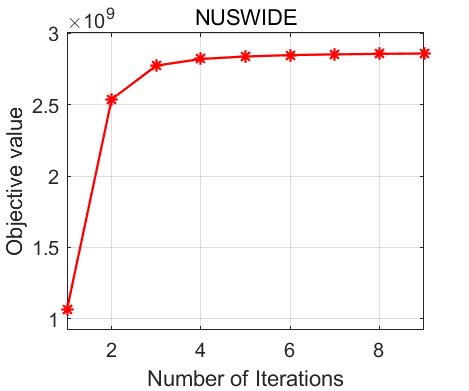}}
		\centerline{(a) Convergence}\medskip
	\end{minipage}
	\hfill
	\begin{minipage}[b]{0.48\linewidth}
		\centering
		\centerline{\includegraphics[width=4.0cm]{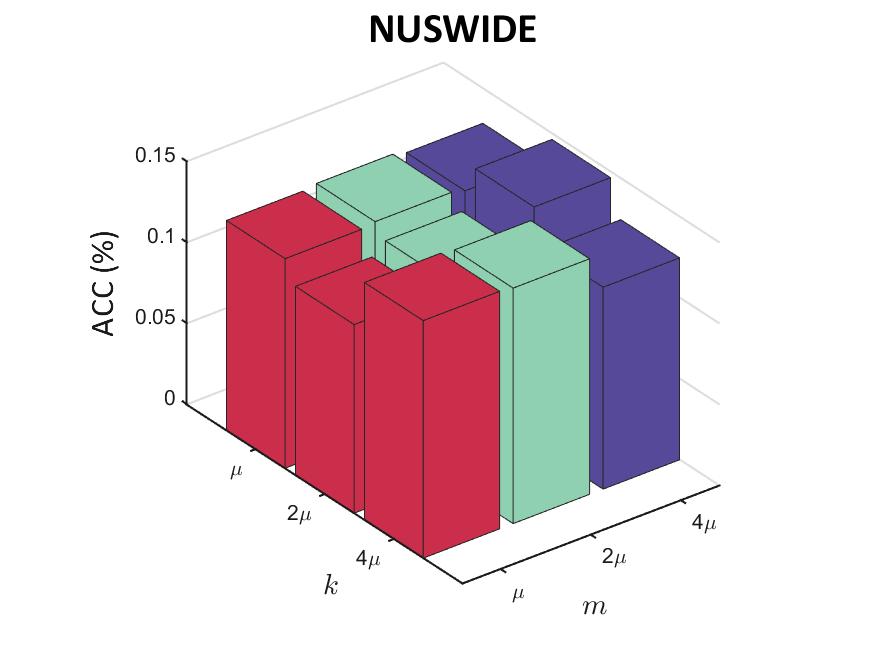}}
		\centerline{(b) Parameter sensitivity}\medskip
	\end{minipage}\vspace{-1mm}
	\caption{Experimental results.}
	\label{fig:res}
\end{figure}

\vspace{-2mm}

\begin{figure*}[htb]
	\centering
	\includegraphics[width=16.9cm]{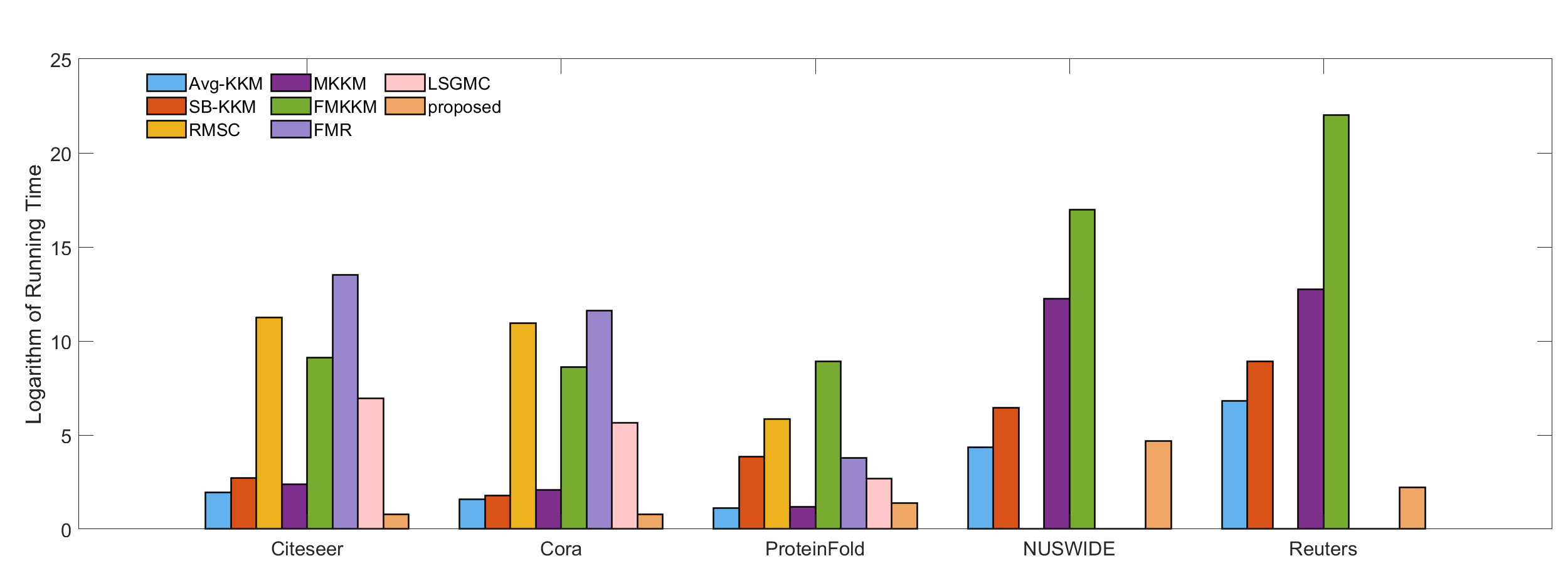}
	\caption{Time comparison.}
	\label{fig:time}
\end{figure*}

\vspace{-2mm}

\noindent\textbf{Convergence} In our six-step iterative optimization process, each updating formula ensures the objective value of the optimization target monotonously increases after each iteration, while keeping the rest five of the decision variables fixed, and our optimization target is upper-bound. We further verify the convergence in Fig.\ref{fig:res}(a).

\noindent\textbf{Parameter sensitivity} We have 2 hyper-parameters in our algorithm: the sampling scale $m$ and number of partitions $k$. Empirical guidance for their pre-set are numerical multipliers of the clustering number $\mu$. Our experiments use $3\times 3$ parameter set $\{\mu,2\mu,4\mu\}$. From Fig.\ref{fig:res}(b), we observe that our method enjoys a stable performance with the variation of hyper-parameters.

\noindent\textbf{Time cost} The running time of compared algorithms on benchmark datasets is shown in Fig. \ref{fig:time}. 
For large-scale datasets (NUSWIDE and Reuters), some algorithms(FMR, RMSC, LSGMC) encounter the `Out-of-Memory' problem. As a result, the time bars of these methods are omitted in the comparison graph.

\section{Conclusion}
In this article, we propose an effective and efficient multi-view clustering method OS-LFMVC-CS, which simultaneously maximizes the alignment between different views and learns the shared subspace, and directly optimizes the cluster assignments. This mechanism greatly enhances the negotiation among views and improves the consistency in shared subspace. In this way, we obtain the clustering results in one step and reduce the time and space expenditure to linear cost. We derive a novel optimization framework using a six-step iterative optimization with verified convergence. In addition, extensive experiments are conducted to support the method well. In the future, we will explore more efficient clustering algorithms and use one-step clustering scheme for easy application on a wide range of scenarios.


\vfill\pagebreak

\bibliographystyle{IEEEbib}
\ninept
\bibliography{refs}

\end{document}